\documentclass{article}

\usepackage{amsmath}
\usepackage{multirow}
\usepackage{graphics}
\usepackage{changes}

\begin{document}
\title{A review on longitudinal data analysis with random forest in precision medicine}

\author{
	Jianchang Hu, 
	Silke Szymczak\thanks{Corresponding author. Institute of Medical Biometry and Statistics, University of Lübeck, Ratzeburger Allee 160, 23562 Lübeck, Germany. Email: silke.szymczak@uni-luebeck.de} \\ 
	\small{Institute of Medical Biometry and Statistics, University of Lübeck, Germany}
}

\date{}
\maketitle

\begin{abstract}
	Precision medicine provides customized treatments to patients based on their characteristics and is a promising approach to improving treatment efficiency. Large scale omics data are useful for patient characterization, but often their measurements change over time, leading to longitudinal data. Random forest is one of the state-of-the-art machine learning methods for building prediction models, and can play a crucial role in precision medicine. In this paper, we review extensions of the standard random forest method for the purpose of longitudinal data analysis. Extension methods are categorized according to the data structures for which they are designed. We consider both univariate and multivariate responses and further categorize the repeated measurements according to whether the time effect is relevant. Information of available software implementations of the reviewed extensions is also given. We conclude with discussions on the limitations of our review and some future research directions.
\end{abstract}


\section{Introduction}\label{sec:intro}

The goal of precision medicine is to provide customized treatments to patients based on their characteristics and thus to improve treatment efficiency while avoiding serious side effects \cite{ashley2016precision,jameson2015precision,matchett2017cancer}. With recent technological advances, large scale genetic and other molecular data can now be collected. Along with demographic and clinical profiles they characterize each patient under different aspects. Typical omics data include gene expression, methylation status, protein or metabolite levels, and microbiome abundances. Many of these measurements, however, change over time, often depending on disease activity, treatment, comorbidities and other environmental factors. Consequently, it is important to measure them for the same patient repeatedly over time, and this leads to longitudinal data, where a single observation captures the measurements at a specific time point for a patient.

Depending on the research question, the study design and the outcome of interest, multiple longitudinal data formats can be envisioned. Predictors might be available for a single time point only, such as at baseline visit, or are time-invariant, which is the case for genetic variants. Alternatively, predictors are measured multiple times during a study; for instance, gene expression or metabolites are measured from multiple blood samples and microbiome abundance is obtained from stool samples collected during a couple of visits.
Similarly, the outcome can be determined at a single time point. Examples include response to treatment at the end of therapy or after a pre-specified follow-up time. But it might also be of interest to predict the outcome over time such as disease activity or severity.
Furthermore, the data format is related to the study design where the same number of measurements at fixed time points is taken for each subject or data from a varying number of irregularly spaced time points are available; the latter is often encountered in observational studies.

In general, a longitudinal data set can be formatted as in Table \ref{tab:str_longitudinal}. Here in total, there are $N$ subjects, for each of them, $n_i$, $i=1, \dots, N$ observations are measured, and each observation consists of measurements on $m$ response variables and $p$ predictors.

\begin{table}[h]
	\centering
	\caption{General structure of longitudinal data}
	\begin{tabular}{c|c|c|c|c|c|c|c}
		\hline
		Subject & Observation/Time & \multicolumn{3}{|c}{Responses} & \multicolumn{3}{|c}{Predictors} \\
		\hline
		1 & 1 & $y_{111}$ & \dots & $y_{11m}$ & $x_{111}$ & \dots & $x_{11p}$ \\
		\hline
		1 & 2 &  $y_{121}$ & \dots & $y_{12m}$ & $x_{121}$ & \dots & $x_{12p}$ \\
		\hline
		$\cdot$ & $\cdot$ & $\cdot$ & $\cdot$ & $\cdot$ & $\cdot$ & $\cdot$ & $\cdot$ \\
		\hline
		1 & $n_1$ &  $y_{1n_11}$ & \dots & $y_{1n_1m}$ & $x_{1n_11}$ & \dots & $x_{1n_1p}$ \\
		\hline
		$\vdots$ & $\vdots$ & $\vdots$ & $\vdots$ & $\vdots$ & $\vdots$ & $\vdots$ & $\vdots$ \\
		\hline
		N & 1 &  $y_{N11}$ & \dots & $y_{N1m}$ & $x_{N11}$ & \dots & $x_{N1p}$ \\
		\hline
		N & 2 &  $y_{N21}$ & \dots & $y_{N2m}$ & $x_{N21}$ & \dots & $x_{N2p}$ \\
		\hline
		$\cdot$ & $\cdot$ & $\cdot$ & $\cdot$ & $\cdot$ & $\cdot$ & $\cdot$ & $\cdot$ \\
		\hline
		N & $n_N$ &  $y_{Nn_N1}$ & \dots & $y_{Nn_Nm}$ & $x_{Nn_11}$ & \dots & $x_{Nn_1p}$ \\
		\hline
	\end{tabular}
	\label{tab:str_longitudinal}
\end{table}

Analyzing longitudinal data is not an easy task. The most distinct feature of longitudinal data is the repeated measurements from the same subject. This inevitably leads to clustered and correlated observations. The clustering effect is due to individual characteristics. For instance, average response to a drug could vary from patient to patient. In the meantime, if repeated measurements are collected over a period of time, then there could be serial correlations among measurements.

Furthermore the observation time for the longitudinal data can be either equally-spaced or irregularly-spaced, which may affect the approach that can be used for the analysis. Visits at every other month would lead to equally-spaced observations, while following up at 6 months, 1 year and 2 years after the treatment provides an example of irregularly-spaced observations. Additionally, irregular spacing can also occur in observational studies when there are no pre-specified follow-up times. One example is the electronic health record data from patient care.
Apart from that, missing values in the repeated measurements are likely to be present. This loss of observations can occur, for instance, when patients are not interested in the follow-up investigations. These missing data pose great challenges to the analysis. They can easily turn an equally-spaced observation schedule into irregular. More importantly, they may carry vital information when the missingness could be related to the value of the variable, which can distort the analysis results if not handled carefully. More discussions on the characteristics of longitudinal data can be found in the classic textbooks \cite{fitzmaurice2012appliedLDA, hedeker2006LDA}.

Despite the difficulties introduced by longitudinal data, they bring rich information. With longitudinal data, clinicians can better understand disease progression, especially of chronic diseases, so that patients can be properly stratified and treatment plans can be tailored accordingly \cite{krasniqidata2021GMDS, Latourelle2017LancetNeurol, zhang2019SciReport}. Furthermore, repeated measurements allow the patients' treatment responses to be captured more accurately, so that effective therapies can be implemented and evaluated.

In order to serve the purpose of precision medicine, the development of prediction models with longitudinal data using statistical or machine learning approaches is crucial \cite{konig_what_2017}. These models, on the one hand, can be applied to predict the current status of an individual, i.e. to evaluate if a specific condition is present (diagnostic setting). On the other hand, they are useful in forecasting if a specific event will occur in the future (prognostic setting). One example would be to predict the future disease course, including the probability of a remission or relapse and the need for therapy changes or intensification.

One of the state-of-the-art machine learning methods for the development of prediction models is the random forest (RF) algorithm \cite{breiman_random_2001, chen_random_2012}. It is a nonparametric approach that can accommodate different types of responses such as categorical or quantitative outcomes and survival times \cite{rfsurvival}. Moreover, it can work with predictors of various scales or distributions and is suited for applications in high-dimensional settings where the number of predictors can be larger than the number of observations \cite{chen_random_2012, cutler_random_2007}. Thus, it is very suitable for analyzing omics data which are often high-dimensional, plus metabolite and protein levels are usually skewed and left censored by limits of detection, and microbiome abundances often exhibit an excess of zeros. Furthermore, tree-based methods form data-driven subgroups of samples which can be beneficial for patient stratification. Via the so-called variable importance measures, the method can also highlight the relevance of each predictor \cite{breiman_random_2001}. This could be especially handy for pharmacogenomics \cite{mooney2015pharmageno, ritchie2012pharmageno}, where potential genetic variants associated with drug response phenotypes such as drug efficacy and adverse side effects can be identified. In fact, Svetnik et al. (2004) demonstrated that the classification and regression tree (CART, \cite{breiman1984classification}) is more powerful in the drug discovery process compared with conventional methods such as partial least squares and support vector machine \cite{svetnik2004CARTapp}.

However, as with other machine learning methods, the RF algorithm assumes that observations are independently sampled from a population. This is unfortunately not the case in longitudinal studies where, as we have pointed out, multiple measurements for the same subject are often collected at different time points. Conducting statistical analysis on longitudinal data without considering the dependency among observations could lead to biased inference due to underestimated standard errors in linear models \cite{raudenbush2002hierarchical} and spurious subgroup identification and inaccurate variable selection in tree-based methods \cite{fokkema2018, sela_re-em_2012}.

Therefore, in this review, we will present a range of extensions of the standard RF algorithm for the analysis of longitudinal data. We limit our attention to CART-based RF with a focus on prediction of categorical and quantitative outcomes. Our review is structured as follows. In the section \ref{sec:uni_response} we consider the case where the response variable is univariate. Here we start with a short review on the standard RF algorithm in subsection \ref{ssec:std_rf}. Following that, in subsection \ref{ssec:uni_no_time} the case of repeated measurements or clustered data is investigated. For such data type, the ordering of the observations by time is ignored in the analysis. Subsection \ref{ssec:uni_with_time} then presents methods that incorporate time effect into modeling. Section \ref{sec:multi_response} focuses on extensions of RF algorithm suitable for multivariate responses. Several extensions are also able to analyze univariate longitudinal data because with suitable transformations the latter can be turned into the multivariate case. After that, we provide information on the currently available implementations of the reviewed methods in section \ref{sec:implementation}. We conclude with a discussion in section \ref{sec:discussion}.

\section{Univariate response longitudinal data}\label{sec:uni_response}

We start with the simple scenario where the response variable is univariate; that is $m=1$ in Table \ref{tab:str_longitudinal}. We first briefly review the standard random forest algorithm as a prediction model, and point out the need for extension in order to better serve the purpose of longitudinal data analysis. Several RF extension methods are then presented and discussed. As we mentioned in the previous section, the most distinct feature of longitudinal data is the repeated measurements, often collected at different time points. Therefore, we categorize these extension methods by their ways of incorporating the time effects.

\subsection{Standard random forest algorithm}\label{ssec:std_rf}

In this section, we give a brief description on the standard random forest algorithm based on CART. More detailed descriptions and discussions on CART and RF can be found in \cite{breiman_random_2001, breiman1984classification, zhang2010treebook}.

Random forest is an ensemble of decision trees where each tree is built from a bootstrapped version of the training data set. Each tree is grown via the principle of repetitive partition where starting from the root node, the same node splitting procedure is applied repetitively until certain stopping rules are met. The main guiding principle for node splitting is to minimize the impurity of response variable in each node of the tree. The impurity of one node is often measured by the Gini index if the response variable is categorical or by the variance if it is quantitative.
For a binary decision tree such as CART, the node splitting process consists of selecting a splitting variable and determining the splitting rule. To choose the splitting variable of a given node, firstly, a subset of predictors are randomly selected. For each predictor $X_i$, splitting rules in the form of $\{ X_i>s \}$ are investigated for all possible cut-off values $s$. The rule $\{ X_i > s^\ast \}$ leading to the maximal impurity reduction of the split becomes a candidate split. Then among all considered predictors, the one with the best candidate split, in terms of the impurity decrease of the split, is selected as the splitting variable and the associated candidate split is the splitting rule of the node.
The growth of each decision tree ends if the nodes to split are already pure (all samples within the node come from the same class or have the same response value) or other pre-determined stopping rules are met (e.g., minimum sample size constrain). The nodes in the final layer of a tree are called leaves and are used for prediction of new observations.

To make prediction with RF, an observation goes through every decision tree in the forest. In each constructed tree, the observation follows the splitting rules and lands in one leaf which predicts its class membership or response value depending on whether it is classification or regression. The final prediction for the observation is made either by majority voting or averaging, based on results from all decision trees in the forest.

Because the RF algorithm uses bootstrap samples to grow each decision tree, some observations are left out in the construction of a given tree. By treating these out-of-bag (OOB) samples as observations needed to be predicted, it can, therefore, provide an estimate of prediction error of the constructed forest.

RF algorithm has been used in many fields including genetic epidemiology, bioinformatics and precision medicine. Its power in prediction comes from the aggregation of many weaker learners. The performance is especially good if the correlations between trees in the forest are low \cite{breiman_random_2001}. In addition, the so-called variable importance measure can be obtained for each predictor, which measures its relevance to prediction. Thus, for high-dimensional dataset such as omics data, variable selection procedures based on variable importance measure are possible (see \cite{silke2019BriefInBioinfo} and the reference therein for a description and comparison of various variable selection procedures based on variable importance measure).

However, one disadvantage of RF is its lack of interpretability. Unlike a single decision tree, the result from a forest is hard to interpret. Even though the variable importance measure can help pinpoint influential predictors, but how these important variables work together is unclear.

\begin{figure}[!h]
	\includegraphics*[width = \textwidth]{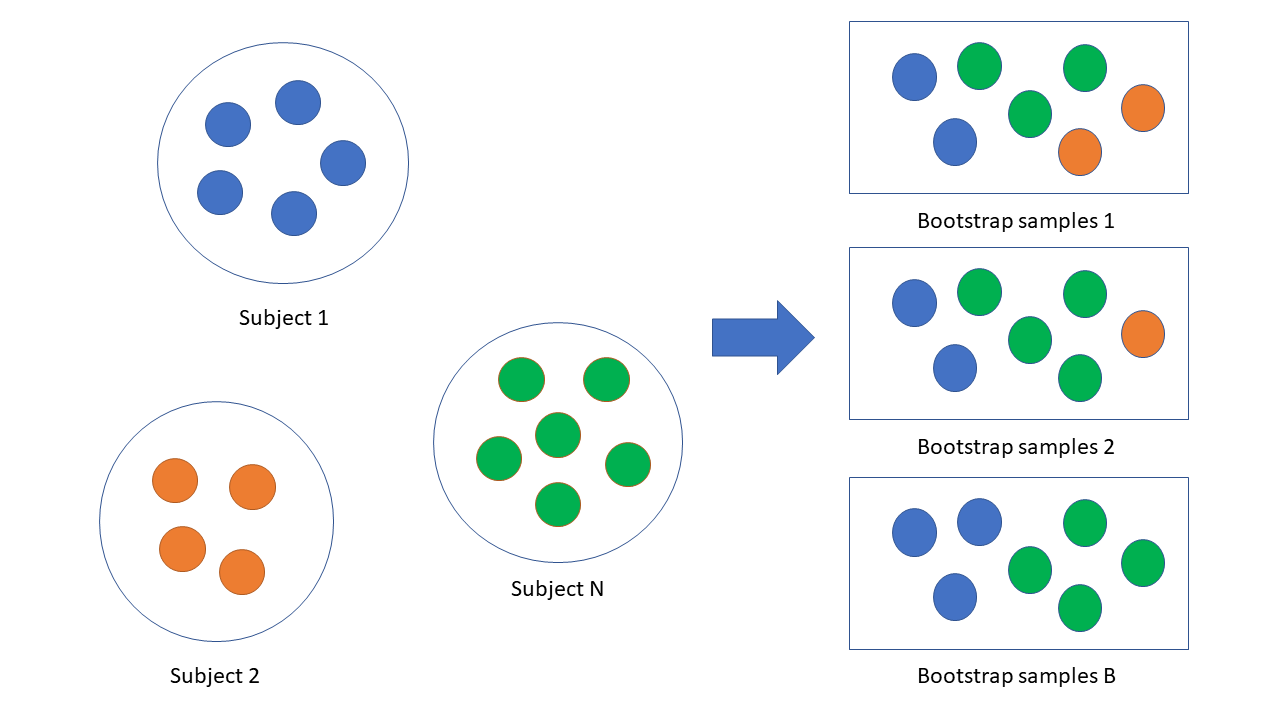}
	\caption{Illustration of bootstrap samples used to construct decision trees in standard RF when it is applied to clustered data.}
	\label{fig:std_rf_cluster}
\end{figure}

Although it is possible to directly utilize the standard random forest algorithm for longitudinal data analysis, it may suffer from several problems. The longitudinal data by nature has a clustered structure. When standard RF algorithm is used directly for analysis, as shown in Figure \ref{fig:std_rf_cluster}, bootstrapped samples may have a high chance to include observations from every subject. This may cause correlated or even homogeneous trees to deteriorate the prediction performance. In addition, the estimated prediction error based on OOB samples is often too optimistic due to the high similarity between the observations from the same subject \cite{karpievitch_introspective_2009}.

Moreover, it has also been reported that ignoring such clustered structure for tree-based methods could result in the detection of spurious subgroups and inaccurate predictor variable selection \cite{martin2015, sela_re-em_2012}. Therefore, there is a need to build extensions of standard RF for longitudinal data analysis. Because a RF is an ensemble of systematically constructed decision trees, the extensions below mainly focus on the modifications of decision tree construction to better fit the longitudinal data.

\subsection{Clustered data}\label{ssec:uni_no_time}

In some applications, the observations are made repetitively on the same subject as duplications. For example, gene expression or metabolites are measured from multiple blood samples and microbiome abundance is obtained from multiple stool samples in a single visit. This results in clustered data setting. In such setting, the data still follow the general format shown in Table \ref{tab:str_longitudinal}, but there is hardly any time effect. In other words, the ordering of the observations from the same subject can be ignored and is not considered in training the prediction model. Hence, one model for clustered data can be written as follows.
\begin{align}\label{eqn:rep_meas}
	y_{ij} = \mu_i + \varepsilon_{ij},
\end{align}
where $\mu_i$ reflects the mean value of subject $i$, and $\varepsilon_{ij}$ are random fluctuations with mean $0$ and independent from each other for all $j=1, \dots, n_i$ and across all $i=1,\dots, N$. The clustering effect, therefore, is the consequence of the shared mean value for observations from the same subject.

\subsubsection{Averaging}\label{sssec:ave_std}

One intuitive approach to deal with the aforementioned clustering effect of repeated measurements is to take the average of replicated data for each subject. This then brings the data structure back to the usual one-subject-one-observation scenario and retain the needed independence for standard RF algorithm. Vlahou et al. (2004) \cite{vlahou2004averageEg} takes this approach to analyze mass spectrometry data for protein profiling in urine.

Despite the simplicity, this approach suffers from a loss of information. The intra-class variation is averaged out. Moreover, this approach also masks the imbalance design. Different subjects could contribute different numbers of observations in the original data set, as in Table \ref{tab:str_longitudinal}, $n_i$ could be different for $i=1,\dots, N$, which may be due to some characteristics of the subjects and may carry underlying distributional information. However, after averaging out the repeated measurements, each subject now makes equal contribution to the training data set. This can have potential effects on the prediction efficiency and variable selection. Karpievitch et al. (2009) \cite{karpievitch_introspective_2009} showed that this approach, when compared with the standard RF, is more sensitive to the total number of subjects $N$; reduction in $N$ leads to poorer prediction and variable selection. The averaging approach may also be difficult to use when classification and categorical predictors are concerned. For a given patient, his/her cholesterol level based on different blood samples may vary which could lead to different categorization; one observation falls into normal level and another belongs to high level. The averaging approach needs to average all observations of this patitent to end up with a subject-level measurement for analysis. However, when different observation-level measurements from the same subject fall in different categories, this averaging would be impossible for categorical variables.

\subsubsection{Subject-level bootstrapping}\label{sssec:subj_boot}

To overcome the disadvantages that averaging approach have, extensions that can utilize all observations are needed. But as we pointed out in Section \ref{ssec:std_rf}, the standard bootstrapping strategy in RF construction can lead to correlated or even homogeneous trees, which devastatingly hurts the prediction performance of standard RF. Also the prediction error based on OOB samples is under-estimated due the similarity between in-bag and out-of-bag data.

\begin{figure}[!h]
	\includegraphics*[width = \textwidth]{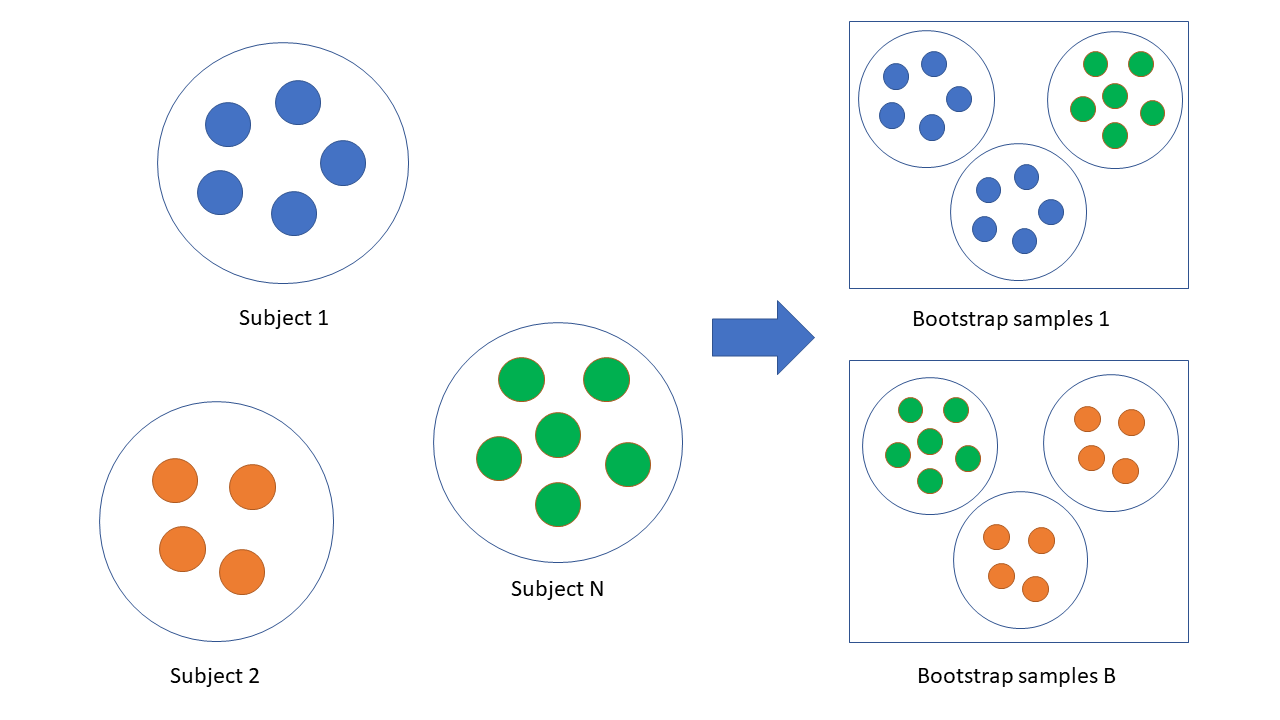}
	\caption{Illustration of subject-level bootstrap samples used to construct decision trees in RF++.}
	\label{fig:subj_bootstrap}
\end{figure}

To tackle these issues while using all observations, Karpievity et al. (2009) \cite{karpievitch_introspective_2009} proposed the subject-level bootstrapping strategy to replace the original one, and the resulting algorithm is named as RF++. Specifically, when building the bootstrap sample to construct a single decision tree in a random forest, instead of re-sampling at the observation level, as shown in Figure \ref{fig:subj_bootstrap}, bootstrap re-sampling at the subject level is performed and all observations from the selected subjects are included as in-bag observations.
Adler et al. (2011a) and Adler et al. (2011b) \cite{adler_ensemble_2011, adler_classification_2011} further extended this idea to a two-stage bootstrapping strategy. Firstly, one subject $i$ is chosen randomly and all associated observations are in bag. Afterwards for each chosen subject the training samples are chosen by randomly selecting one observation from all $n_i$. Adler et al. (2011b) \cite{adler_classification_2011} showed in their simulation studies that subject-level resampling based on one observation per subject yields the best prediction results compared to the standard RF, averaging approach and RF++, although one should also notice that different settings may lead to different results and there could be cases where the other methods are more preferable.

The adoption of subject-level bootstrapping avoids the problem of potentially exposing individual trees to all subjects. The two-stage bootstrapping strategy could further mitigate the negative effect the intra-cluster correlation casts on the prediction performance; when only one observation per subject is selected, even though the same subject might be used in construction of different trees, likely different observations are selected for the training of different trees, which further reduces the similarity between trees.

Besides the usual observation-level classification, Karpievitch et al. (2009) \cite{karpievitch_introspective_2009} showed that classification at subject level is also possible. A majority vote can be performed across the observations belonging to the same subject to result in the subject classification. With such results, a subject-level misclassification rate estimate based on OOB samples is also made possible. This information may be more beneficial and easier to interpret in clinical trials.

However, as pointed out in Hajjem et al. (2014) \cite{hajjem_mixed-effects_2014}, the subject-level bootstrapping only adjusts the sampling method for clustering, thus, no random effects are incorporated in the modeling as well as prediction. Furthermore, for longitudinal studies where time plays a role, this strategy cannot fully utilize the information contained in the data set.

\subsection{Time effects considered}\label{ssec:uni_with_time}

For many research questions, not only values of predictors at the current time point, but also from the past are helpful, sometimes even crucial, for a good prediction performance. A large value of a particular biomarker might be relevant if it had rather small values in the past, pointing to an early change on the molecular level. Therefore, in this section, we would like to review several RF extensions that take time effects into consideration.

\subsubsection{Historical RF}\label{sssec:htree}

The historical RF is an approach that explicitly considers the history of predictors \cite{htreeManuscript}. Assume that we have training data $\{y_{ij} , t_{ij} , X_{ij}\}$, $i = 1, \dots, N$ and $j = 1, \dots, n_i$. Here $y_{ij}$ denotes the response, $X_{ij}$ the vector of predictors and $t_{ij}$ the time of the $j$-th observation on the $i$-th subject. The method estimates a model for the response $y_{ij}$ using both $(t_{ij} ; X_{ij})$ (the observations concurrent with $y_{ij}$) and all preceeding observations of the $i$-th subject up to (but not including) time $t_{ij}$. Thus, for a time-varying predictor, its historical information along with its current value are both used for modeling. For a time-invariant predictor, of course, only its current value is used as in the standard RF.

For time-invariant predictors the standard splitting procedure described in Section \ref{ssec:std_rf} is adopted when constructing each decision tree. In case of a time-varying predictor, its historical information, i.e., values within a specific time interval before the time concurrent with $y_{ij}$, is first represented by a summary function. One exemplary such function for subject $i$ at time point $j$ counts the number of past observation values, including both response and predictor variables, that are measured at a maximum of $\eta_1$ units of time before the current time point $j$ and smaller than $\eta_2$, i.e.,
\begin{align}\label{eqn:htree_summary_fun}
	s_c(\eta;\bar{z}_{ijk}) = \sum_{t_{il}\in [t_{ij}-\eta_1, t_{ij})} I(z_{ilk} < \eta_2),
\end{align}
where $\bar{z}_{ij} = \{ z_{il}=(y_{il}, x_{il}): t_{il} < t_{ij} \}$ denotes the past observations of subject $i$ prior time $j$, and $\bar{z}_{ijk}$ is its $k$-th component. This aggregation results in a single number per observation and variable for a fixed value of $\eta = (\eta_1, \eta_2)$. Alternative summary functions are the relative frequency of values above a certain threshold or the mean. For each summary function, there is also a windowed versions where the time interval considered is further limited by an upper bound, i.e., $t_{il}\in [t_{ij}-\eta_1, t_{ij}-\eta_3)$ in equation \eqref{eqn:htree_summary_fun}. The different functions usually lead to similar prediction performance (personal communication). However, it should be noted that only the frequency based functions are scale invariant which is one of the properties that make the standard RF algorithm robust. Finally, the partitioning at a particular node is performed using the predictor with the smallest Gini impurity or sum-of-squares error for categorical or quantitative response, respectively. However, determination of an optimal cut-off point for time-varying predictors includes optimization of the parameters in the summary function such as $\eta$ which largely increases the computing expenses especially when the number of time-varying predictors is large such as in some omics datasets.

To mitigate the effects of addition optimization of the parameters in the summary function, Sexton and Laake \cite{htreeManuscript, htreeR} incorporates an additional level of randomization where instead of using all observations within the specified time interval, only a sub-sample is randomly selected and used for optimizing the cut-off point. In addition, subject-level re-sampling strategy is also adopted in random forest construction. This not only enjoys the advantages mentioned in Section \ref{sssec:subj_boot}, but also keeps the complete observation history of a subject.

As variable importance measure, Sexton and Laake \cite{htreeManuscript} consider a delete-variable approach. Specifically, to find the importance measure of predictor $k$, the prediction errors of historical RF models with and without this predictor are calculated, and their difference gives the importance measure of the predictor. This approach is computationally demanding. Furthermore, correlation among predictor variables could affect the importance measure since masking may make important variables seem not to be important at all.

\subsubsection{Extensions from (generalized) linear mixed effects model}

A different approach to adjust for the longitudianl structure is to combine (generalized) linear mixed models ((G)LMMs) with the decision tree or RF algorithm. The (G)LMM is a classic statistical methodology for the analysis of longitudinal or more general clustered data. As with (G)LMMs the predictors can be constant or varying over time and different time points are possible for each subject. One advantage of (G)LMM is its explicit modeling of intra-subject correlation structure as well as subject-level random effects besides the main fixed effects of interest. After properly adjusting for these random effects and correlation structure, the longitudinal data becomes conditionally independent, thus the estimation of the fixed effect component of the model follows exactly the same way as if independent observations were observed. Moreover, prediction can now be generalized to a wider population. However, the drawbacks of this approach include its computational complexity to fit mixed effects models as well as the possibility to misspecify the intra-subject correlation structure. More detailed descriptions, discussions and applications on classical methods for longitudinal data analysis can be found in several textbooks, e.g. \cite{fitzmaurice2012appliedLDA, hedeker2006LDA}.

The general idea of the RF extension from (G)LMM is to replace the linear model of the fixed effect component by a tree or RF while keeping the modeling of the dependence structure with random effects. Multiple algorithms have been developed to incorporate the tree or RF into the (G)LMM and are summarized in Table \ref{tab:ext_glmm}. As can be seen, most extensions are based on two approaches, namely, MERT and RE-EM trees. Also for binary response, a Bayesian approach called BiMM has been proposed.

\begin{table}[h]
	\centering
	\caption{Overview of different RF extensions from (G)LMM.}
	\begin{tabular}{l l l}
		\hline
		Outcome & Tree & Forest \\
		\hline
		Quantitative (Gaussian) & MERT & MERF \\
		 & RE-EM tree & REEMforest \\
		 & SMERT & SMERF \\
		 & SREEM tree & SREEMforest \\
		Exponential family & GMERT &  \\
		 & GMET & GMERF \\
		Binary & BiMM tree & BiMM forest \\
		\hline
	\end{tabular}
	\label{tab:ext_glmm}
\end{table}

We first describe approaches for a regression setting based on LMMs, followed by more general methods using GLMMs that can be employed in the context of classification but also for other types of outcomes such as count variables.
\newline

\noindent\emph{Quantitive (Gaussian) response variable}
\newline

\noindent For a normally distributed quantitative outcome, the classic LMM model can be written as
\begin{align*}
	y_i &= X_i\beta + Z_ib_i + \epsilon_i, \\
	b_i &\sim N(0, D), \epsilon_i\sim N(0, R_i),
\end{align*}
where $y_i = (y_{i1}, \dots, y_{in_i})'$ is the $n_i \times 1$ vector of the outcome for the $n_i$ observations of subject $i$, $X_i = [x_{i1}, \dots, x_{in_i}]'$ is the $n_i \times p$ matrix of predictors considered as fixed effects, $Z_i = [z_{i1}, \dots, z_{in_i}]'$ is the $n_i \times q$ matrix of predictors modeled as random effects, $\epsilon_i = (\epsilon_{i1}, \dots, \epsilon_{in_i})'$ is the $n_i \times 1$ vector of random errors, $\beta$ is the $p \times 1$ unknown vector of parameters of the fixed effects, and $b_i$ is the $q \times 1$ unknown vector of random effects of subject $i$. Both $b_i$ and $\epsilon_i$ are assumed to follow a normal distribution with mean zero and covariance matrix $D$ and $R_i$, respectively. It is further assumed that they are independent and that the observations between subjects are also independent. The parameters can be estimated by maximum likelihood (ML) or restricted maximum likelihood (REML) methods.

Two different strategies have been proposed in the literature to replace the linear dependency between the predictors and the outcome. The first approach is the mixed effects regression tree/forest (MERT \cite{hajjem_mixed_2011} and MERF \cite{hajjem_mixed-effects_2014}) where the fixed effects are estimated using a standard regression tree or RF. Specifically, the modified model can be written as
\begin{align}\label{eqn:lmm_tree}
	y_i &= f(X_i) + Z_ib_i + \epsilon_i, \\
	b_i &\sim N(0, D), \epsilon_i\sim N(0, R_i), \nonumber
\end{align}
where it is further assumed that $R_i = \sigma^2 I_{n_i}$ where $I_n$ denotes the identity matrix with size $n$, and the function $f(X_i)$ is estimated by the standard tree or RF. For model fitting an expectation-maximization (EM) algorithm \cite{laird1982random} is used which iterates between estimation of the fixed and random effects components. The general approach can be described as follows (slightly modified from \cite{hajjem_mixed_2011} and \cite{hajjem_mixed-effects_2014}):
\begin{enumerate}\label{algorithm:em}
	\item Initialize with $\hat{b}_i=0$, $\hat{\sigma}^2 = 1$, and $\hat{D} = I_q$;
	
	\item Iterate through the following steps until convergence:
		\begin{enumerate}
			\item[(a)] Estimate a regression tree or RF based on the new outcome variable $y_i^\ast = y_i - Z_i\hat{b}_i$, and predictors $X_i$. Denote the predictions with $\hat{f}(X_i)$;
			
			\item[(b)] Fit the linear random effect model $y_i = \hat{f}(X_i) + Z_ib_i + \epsilon_i$.
		\end{enumerate}
\end{enumerate}
The convergence is based on a generalized log-likelihood criterion
\begin{align*}
	GLL(f,b_i|y) &= \sum_{i=1}^N\{ [y_i-f(X_i)-Z_ib_i]^TR_i^{-1}[y_i-f(X_i)-Z_ib_i] \\
				 &+ b_i^TD^{-1}b_i + \log|D| + \log|R_i| \}.
\end{align*}

This method assumes that the correlation is only due to between subject variation, i.e. the covariance matrix $R_i$ of the errors $\epsilon_i$ is assumed to be diagonal. The MERT approach uses a decision tree to estimate $f(X_i)$, while the MERF method improves prediction performance by considering a standard random forest. It can be noticed, that in MERF the bootstrap sample for each tree is drawn on the observation level and predictions are based on the out-of-bag sample to reduce the risk of overfitting. Note that resampling of individual observations is possible in this setting since it is assumed that the correlation between observations can be completely modeled by the random effects. Thus, using the modified outcome variable $y_i^\ast$ results in independent observations.

The second approach was independently proposed in \cite{sela_re-em_2012} and is called random effects expectation-maximization (RE-EM) trees. It still considers the model \eqref{eqn:lmm_tree}, but it does not directly use tree or RF algorithms to estimate the fixed effects. Instead it considers the partition of samples formed by the regression tree and estimates local fixed effects within each partition while  estimating the random effects globally. The algorithm is similar to MERT in using the generalized log-likelihood as convergence criterion.

More specifically, for model fitting, step 1 is the same as for MERT. Step 2 is modified as follows.
\begin{enumerate}\label{algorithm:reem_step2}
	\item[2.] Iterate through the following steps until convergence:
	\begin{enumerate}
		\item[(a)] Estimate a regression tree or RF based on the new outcome variable $y_i^\ast = y_i - Z_i\hat{b}_i$, and predictors $X_i$. Construct indicator matrix $\Phi^i$ with size $n_i\times T$ where $\Phi_{jt}^i = I(y_{ij}\in g_t)$, $I(\cdot)$ is the indicator function and $g_t$ is the $t$-th terminal node of the tree, and $T$ is the total number of terminal nodes;
		
		\item[(b)] Fit the linear mixed effects model
		\[ y_i = \Phi^i\mu + Z_ib_i + \epsilon_i, \]
		where $\mu = (\mu_1, \dots, \mu_T)$ denotes the local fixed effects within each terminal node.
	\end{enumerate}
\end{enumerate}

The tree is thus only used to define the partition of the sample space and a system of LMM models is fitted with global random effects and each partition having its own local fixed effects. The \emph{lme} function in the R package \emph{nlme} is employed for LMM model fitting which allows a general within-subject correlation structure; for instance, an autocorrelation structure within the errors is possible so that $R_i$ can be a non-diagonal matrix.

An extension of the RE-EM tree is called REEMforest where an ensemble of RE-EM trees is generated for the fixed effects estimation \cite{capitaine_2021}. The function $\hat{f}(X_i)$ is estimated by the mean of the $K$ fitted RE-EM trees:
\[ \hat{f}(X_i) = \frac{1}{K}\sum_{k=1}^{K}\Phi^{i,k}\hat{\mu}_k, \]
where $\Phi^{i,k}$ is the $n_i\times T$ indicator matrix based on the tree $k$ and $\hat{\mu}_k$ is the $T\times 1$ vector of fitted local fixed effects from tree $k$.

One prominent feature of longitudinal data is its serial correlation within the observations of the same subject. In order to model such a covariance structure that varies over time, the MERT and RE-EM tree and their corresponding forest variants have also been extended to include an additional stochastic component \cite{capitaine_2021}. The resulting approaches are correspondingly called SMERT, SREEMtree etc. The model with the additional stochastic component can be written as follows:
\[ y_i = f(X_i) + Z_ib_i + \omega_i + \epsilon_i, \]
where $\omega_i = (\omega_i(t_1), \dots, \omega_i(t_{n_i}))'$ is a centered Gaussian process with $\text{Cov}(\omega_i(s), \omega_i(t)) = \gamma^2\Gamma(s, t)$. The $\omega_i(t)$ are independent for different subjects $i=1, \dots, N$ and $b_i,\ \epsilon_i$ and $\omega_i(t)$ are mutually independent. Again, a variant of the EM algorithm is used to estimate the parameters where the definition of the new variable $y_i^\ast$ now also includes the additional stochastic component: $y_i^\ast = y_i - Z_i\hat{b}_i - \hat{\omega}_i$. In their simulation studies, Capitaine et al. (2021) \cite{capitaine_2021} showed that both MERT and RE-EM based tree and RF algorithms are applicable to high-dimensional datasets. Furthermore, they demonstrated that tree- and forest-based extensions provide more accurate prediction than LMM and standard RF. For extensions with stochastic processes, misspecification, where true underlying data generating mechanism uses either no stochastic process or other processes than the one adopted in the estimation procedure, has only limited impact on prediction performance. Besides, variable selection via variable importance measure is possible for these methods. These characteristics make the extensions suitable for omics data analysis. In fact, \cite{capitaine_2021} compared all forest-based extensions, i.e., MERF, REEMforest, SMERF and SREEMforest on the DALIA vaccine trial dataset where expression of 32,979 gene transcripts was included in the analysis.

Before we move on to the context of generalized linear models, we would like to remark that for predicting the outcome of new observations with aforementioned RF extensions, two different settings have to be distinguished. The first case is prediction for a new subject $i$ for which no random effects $\hat{b}_i$ are available. Thus, prediction is solely based on the fixed effect component which is either given by the prediction of the tree or RF ($\hat{f}(X_{it})$) or the predicted effect associated with the terminal node in which the new observation lands ($\Phi_{jt}^i\hat{\mu}_t$). Secondly, to predict a new observation for a subject $i$ used in the training process, the sum of the fixed component and the corresponding random effect of subject $i$ can be used.
\newline

\noindent\emph{Generalized response variable}
\newline

\noindent The approaches described so far in this section assume a quantitative outcome that is normally distributed. Further extensions have been proposed for other types of outcomes by using generalized linear mixed models (GLMMs)  instead of LMMs.

The GLMM assumes that, conditional on the random effects, the outcome $y_i$ follows a distribution from the exponential family. The GLMM model can be further specified as:
\begin{align*}
	g(\mu_i) &= \eta_i = X_i\beta + Z_ib_i, \\
	b_i &\sim N(0, D),	
\end{align*}
where $\mu_i = E(y_i|b_i)$, $g(\cdot)$ is a known link function, and $\eta_i$ is a $n_i \times 1$ vector. The commonly used link functions include identity link, logit link and log link functions for quantitative, binary and count outcomes, respectively. Parameters of GLMMs are estimated by ML or REML methods using numerical optimization algorithms such as penalized quasi-likelihood (PQL) \cite{rodriguez2008multilevel}, iteratively reweighted least squares or a Newton-Raphson method \cite{mccullagh2019generalized}.

Similar to the quantitative outcome case, the RF extensions from GLMM replace the linear relationship between outcome and fixed effects predictor variables by a nonparametric alternative such as a decision tree or RF. The essential estimation procedure is again using an iterative algorithm inspired by the EM algorithm \cite{laird1982random} to estimate the fixed and random effects separately and iteratively. Here, we only provide a brief summary of the approaches and mention their quantitative counterparts. For more details, we refer the reader to the original publications.

The MERT approach has been extended to the generalized mixed effects regression tree (GMERT) \cite{hajjem_generalized_2017}. The PQL algorithm of the GLMM is modified so that a weighted MERT pseudo-model is used instead of the weighted linear mixed-effects pseudo-model. The fixed part is again estimated with a standard regression tree. In this implementation it is necessary to specify initial estimates of the mean values $\hat{\mu}_i$. In the simulation study with a binary outcome the authors used pre-determined values $\hat{\mu}_{ij} = 0.25$ if $y_{ij} = 0$ and $\hat{\mu}_{ij} = 0.75$ if $y_{ij} = 1$. Unfortunately, no further discussions on this initialization were presented.

Similarly, the generalized mixed effects tree (GMET) extends the RE-EM tree \cite{fontana2018performing}. Again, a regression tree is used and the indicator variables for the terminal nodes are modeled as fixed effects in the mixed effects model. The modified outcome variable for the regression tree is $y_i^\ast = \eta_i - Z_ib_i$. However, $\eta_i$ needs to be estimated which is usually achieved with a standard generalized linear model (GLM) using the predictors as fixed effects covariates (in \cite{fontana2018performing, pellagatti_generalized_2021}). Note that this approach is not possible for high dimensional data due to this need to estimate $\eta_i$ with GLM since the number of variables then cannot exceed the number of observations. To the best of our knowledge, no solutions for high dimensional data have so far been proposed in the literature along this direction. GMET has further been extended to generalized mixed effects random forest (GMERF) \cite{pellagatti_generalized_2021} where instead of growing only a single decision tree, a random forest is trained.

\subsubsection{A Bayesian approach}

For binary outcomes, Binary Mixed Model (BiMM) tree \cite{speiser_bimm_2020} considers a Bayesian implementation of GLMM. The GLMM portion of the BiMM method has the form
\[ \text{logit}(\mu_{it}) = \text{CART}(X_{it})\beta + Z_{it}b_{it}, \]
where $\text{CART}(X_{it})$ are indicator variables reflecting membership of each longitudinal observation $t$ for subject $i$ in terminal nodes within the decision tree. Therefore, the use of the tree in this approach is again not to model the fixed effects directly, but rather to determine similar groups of observations after random effects have been properly adjusted.

For estimation the BiMM tree method again adopts the EM-like algorithm and iterates between developing CART models using all predictors and then using information from the CART model within a Bayesian GLMM to adjust for the clustered structure of the outcome. Specifically, the procedure can be briefly summarized as follows.
\begin{enumerate}
	\item[(a)] CART construction with $(y_i, X_i)$ and obtain predicted probability $p_\text{CART}(X_{it})$
	
	\item[(b)] fit Bayesian GLMM with $(y_{it}, p_\text{CART}(X_{it}), Z_{it})$
	
	\item[(c)] update $y_{it}^\ast$ by discretization of $y_{it} + q_{it}$ where $q_{it}$ is the predicted probability from the Bayesian GLMM
	
	\item[(d)] repeat (a)-(c) with $y_{it}^\ast$ until the change in posterior log-likelihood is less than a specified threshold
\end{enumerate}

This tree method is further extended to a forest-based method where all $\text{CART}(X_{it})$ are replaced by $\text{RF}(X_{it})$. More details of the algorithms can be found in \cite{speiser_bimm_2019, speiser_bimm_2020}.

Compared with the previously reviewed frequentist methods, the Bayesian approach, as pointed out by the authors, can avoid issues with model convergence, especially when data are high dimensional. In addition, when uninformative priors are used, frequentist GLMM results can be obtained. That is to say, the Bayesian approach provides a more general framework with frequentist approaches such as RE-EM tree/forest as special cases.

\section{Multivariate response longitudinal data}\label{sec:multi_response}

So far, the reviewed methods are designed for univariate response variables, however, often an array of health-related symptoms or scores could be of interest at the same time, which leads to multivariate responses. Moreoever, even with an univariate response, we can also treat measurements at different time points together as multivariate responses or a discretized response curve. Therefore, in this section, we would like to shift our attention to extensions of the RF algorithm that can accomodate multivariate response variables.

Before we start, we would like to note that the algorithms in this section are directly applicable with time-invariant predictors such as genetic data. If predictors are also observed at multiple time points, techniques from the previous sections, such as subject-level bootstrapping, historical RF, and incorporation of mixed effects etc, need to be used along with the modifications reviewed in this section for an adequate analysis.

\subsection{Repeatedly measured univariate longitudinal responses as multivariate response}\label{ssec:multi_rf}

One distinct feature of longitudinal data is the repeated measurements at different time points, which leads to dependence between observations. However, at the subject level, the usual independency assumption is still reasonable. Therefore, one strategy to analyze longitudinal data is to consider observations at different time points jointly so that each subject has only one multi-dimensional response. But within the multi-dimensional response, variables at different dimensions are not independent as they represent the same measurement taken at different time points. Therefore, when this multivariate response approach is considered, on the one side, we have the independency between samples, but on the other side, we still need to take care of the inter-dimensional correlation.

To accomodate multivariate responses, the common strategy to extend the RF algorithm largely focuses on modifying the split criterion in the construction of each decision tree, where impurity measures are modified so that multivariate responses can be handled properly. In addition the covariance structure needs to be considered when defining the impurity measure to account for the inter-dimensional correlation. The modifications can be roughly categorized into two classes, using either distance or likelihood based split criteria.

\subsubsection{Distance based impurity functions}\label{sssec:dist_impurity}

Segal (1992) \cite{segal_tree-structured_1992} was among the first to extend CART to longitudinal data by using a distance based measure for node impurity. Specifically, the author considered an univariate quantitative outcome but treated measurements at different time points jointly as a multivariate response. It is further assumed that the observation times for all subjects are the same, so that the dimension of the multivariate response is fixed and not changing across subjects. For a given node $t$, Segal (1992) \cite{segal_tree-structured_1992} considers the following generalized sum of square function:
\begin{align}
	\label{eqn:gss}
	SS(t) = \sum_{i\in t} (y_i-\bar{y}_t)'\mathbf{V}(\theta, t)^{-1}(y_i-\bar{y}_t),
\end{align}
where $y_i,\ i = 1\dots, N$ is a $n\times 1$ vector, $\bar{y}_t$ is the sample average of $y_i$'s within node $t$,  $\mathbf{V}(\theta, t)$ denotes the $n\times n$ covariance matrix of the responses within node $t$ and depends on unknown parameters $\theta$ which can be estimated within the node. Then the splitting rule $s$ of the node $t$ is evaluated via
\[ \phi(s, t) = SS(t) - SS(t_L) - SS(t_R). \]
In principle, the estimated parameters $\hat\theta$ can differ for node $t$ and its daughters $t_L$ and $t_R$, which as the author noticed may lead to negative $\phi$. Hence, the author further imposes the restriction that for each candidate split the covariance parameters are determined from the parent node $t$ so that
\[ \mathbf{V}(\theta, t) = \mathbf{V}(\theta_L, t_L) = \mathbf{V}(\theta_R, t_R). \]
Furthermore, the author provides several candidates for the covariance structure, namely, independence (i.e., diagonal matrix), first-order autoregression (AR1), compound symmetry (CS), and sample covariance matrix.

The independence structure leads to the sum of square about the mean:
\[ SS(t) = \sum_{i\in t}\sum_{j=1}^n (y_{ij}-\bar{y}_j)^2, \]
where $y_{ij}$ is the outcome for subject $i$ and component $j$, and all subjects are assumed to have same number of $n$ components. This is a direct generalization from the univariate regression tree, and has been used by De’Ath (2002) \cite{death_multivariate_2002} for applications in ecology and by Segal and Xiao (2011) \cite{segal_multivariate_2011} in the construction of the multivariate random forest.

When the sample covariance matrix is adopted in Segal's approach, the generalized sum of square function is closely related to the Mahalanobis distance where the Mahalanobis distance of an observation $y_i$ from a set of observations with mean $\mu_i$ and (nonsingular) covariance matrix $\mathbf{S}$ is defined as
\[ D(y_i) = \sqrt{(y_i-\mu_i)'\mathbf{S}^{-1}(y_i-\mu_i)}. \]
Larsen and Speckman (2004) \cite{larsen_multivariate_2004} directly considered the Mahalanobis distance as node impurity measurement and split criterion. Instead of updating the covariance structure during the tree construction, they estimate the covariance matrix from the whole data set at the very beginning and use the estimate throughout the whole process. They still consider the simple average of observations in each node for $\mu_i$, but different estimators such as trimmed mean could also be adopted.

Besides the Mahalanobis distance based split criterion, De'Ath (2002) \cite{death_multivariate_2002} proposed the distance-based multivariate regression tree (db-MRT), where the impurity of a given node is measured based on the pair-wise dissimilarities between observations within the node. Sim et al. (2013) \cite{sim_random_2013} put this approach into a more formal construction where the dissimilarities between observations are captured by a distance matrix. Here, a distance matrix $D$ is a symmetric positive real-valued matrix, where the components $D_{ij}$ denote the distances between $y_i$ and $y_j$ and $D$ satisfies the three required distance conditions $D_{ii}=0, D_{ij}=D_{ji}, \text{ and } D_{ij} + D_{jk} \geq D_{ik}, \forall i, j, k=1, \dots, N$. Then the impurity of each node $t$ is defined as
\[ Imp(t) = \sum_{i,j\in t} D_{ij}^2, \]
The split criteria $s$ is evaluated via
\[ \phi(s) = -(Imp(t) - Imp(t_L) - Imp(t_R)), \]
and the one achieving the maxium gives the optimal splitting criteria.

This approach is more general than the aforementioned extensions in that the distance matrix does not necessarily depend on the dimension of the original responses. In fact, it is possible to analyze longitudinal responses at irregular time points with this approach as long as an appropriate distance measure can be defined. However, how to make prediction with the resulting RF needs further consideration because now within a leaf, it is possible to have responses with different dimensions, thus usual sample average would not make sense in such cases. This approach may also be applicable for analyzing multiple longitudinal responses. As long as the distance matrix between pairs of responses can be properly defined, then the construction of the tree and RF does not depend on the dimension of the responses. 

Lastly, when the distance is measured by $l_1$-norm, given the well-known relationship that
\[ \sum_{i=1}^N|y_i - \bar{y}|^2 = \frac{1}{2N}\sum_{i=1}^N\sum_{j=1}^N |y_i - y_j|^2, \]
this distance matrix based approach is connected to Segal's approach with an assumed independence covariance structure.

\subsubsection{Likelihood based impurity function}\label{sssec:likeli_impurity}

Zhang (1998) \cite{zhang1998classification} extended CART to multiple binary response variables. For responses from an exponential family distribution, the author considered the log-likelihood as the node impurity that depends only on the linear terms and the sum of the second-order products of the responses. Specifically, for subject $i$, $y_i$ is assumed to follow the joint probability distribution:
\[ f(y_i; \Psi, \theta) = \exp(\Psi'y_i + \theta w_i - A(\Psi, \theta)), \]
where $\Psi$ and $\theta$ are arrays of parameters, $A(\Psi, \theta)$ is the normalization function depending only on $\Psi$ and $\theta$, and $w_i = \sum_{j<k} y_{ij}y_{ik}$. The node impurity is defined as the maximum of the log-likelihood derived from this distribution; that is, for node t,
\[ Imp(t) = \sum_{i\in t} (\hat{\Psi}'y_i + \hat{\theta}w_i - A(\hat{\Psi}, \hat{\theta})), \]
where $\hat{\Psi}$ and $\hat{\theta}$ are the maximum likelihood estimates of $\Psi$ and $\theta$ within the node.
Zhang and Ye (2008) \cite{zhang2008ordinal} applied the same technique to ordinal responses by first transforming them to binary-valued indicator functions.

When multivariate normally distributed responses are considered, Abdolell et al. (2002) \cite{abdolell2002} proposed a likelihood-ratio test statistic as impurity function. Specifically, suppose that $y_i\sim N_n(\mu_i, \Sigma)$, the authors define the deviance function for a single observation as
\[ D(\mu_i; y_i) = 2[\ell(y_i;y_i) - \ell(\mu_i; y_i)] = (y_i-\mu_i)'\Sigma^{-1}(y_i-\mu_i), \]
where $\ell(\mu_i;y_i)$ is the log-likelihood function. Assuming that $\Sigma$ is constant and given for all $i$, they further define the deviance within a node $t$ as
\[ D(\hat{\mu}; y, t) = \sum_{i\in t} D(\hat{\mu}; y_i), \]
where $\hat{\mu}$ is the restricted maximum likelihood estimate (REML) within the node. The impurity of a node is then measured by the negation of the deviance. As pointed out by the authors, this deviance function in the context of the multivariate normal distribution, is the Mahalanobis distance between $\mu_i$ and $y_i$. In addition, they also noticed that deviance assessed via the multivariate analysis of variance (MANOVA) approach such as Hotelling's $T^2$ is again in a form of the Mahalanobis distance. These observations connects the likelihood-ratio test statistics based impurity function with the aforementioned Mahalanobis distance based one.

The likelihood-ratio test statistics based impurity function is also considered by Segal (1992) \cite{segal_tree-structured_1992} for multivariate normally distributed responses. However, their splitting rule focuses on the intra-cluster variation structure of subjects other than the mean structure of responses, which we think may be difficult to interpret and less of interest in terms of precision medicine. 

\subsection{Multiple longitudinal responses}\label{ssec:uni_to_multi}

If an array of health-related symptoms are monitered at the same time, then this leads to multiple longitudinal responses. The extensions in the previous section, except the db-MRT, may not be directly applicable because if the repeated measurements at different time points are also considered together, then for each subject $i$, its corresponding reponse $y_i$ is a matrix (see Table \ref{tab:str_longitudinal}).

One possible approach is to combine the extensions in the previous section with the techniques we reviewed in Section \ref{sec:uni_response} such as subject-level bootstrapping and mixed effects models. Another approach is given by Yu and Lambert (1999) \cite{yu1999functional} which can be considered as an extension from the traditional non-parametric and semi-parametric regression models for longitudinal data (see Chapter 8-12 of \cite{fitzmaurice2008LDAbook} for detailed review and discussions on traditional methods). For longitudinal data observed at many observation times, they treat each response vector as a random function and fit each trajectory with a spline curve. Then they use the estimated coefficients of the basis functions as multivariate responses to fit a regression tree model. When the number of observation times is large, this approach can effectively reduce the dimensionality of the responses. Furthermore, by considering the same set of expension basis for all trajectories, the data structure is unified, so this approach is applicable to irregular-spaced observations.

\section{Implementation}\label{sec:implementation}

In Table \ref{tab:implementation}, we provide a summary of the software implementations of the RF extensions reviewed in previous sections. For each method, its implementation and where to obtain the package or code is listed. Information on the type of problem the extension method can solve is also given and whether the method grows a tree or random forest is summarized. If variable importance measure is supplied by the package or code, such information is also presented.

Briefly, almost all extensions are presented in an R package on CRAN or as an R program. Programs of MERT and MERF can only be obtained directly from their authors, no public access is available. The majority of extensions do not supply variable importance measures, which could be a future research direction for longitudinal data analysis with random forests. We also remark that no systematic comparisons and efficiency studies have been performed so far on the listed packages and programs.

\begin{table}[h]
	\caption{Overview of the implementations of the reviewed RF extensions.}
\resizebox{\columnwidth}{!}{
	\begin{tabular}{l l l l l l }
		\hline
		Name & Implementation & Type & Response & VImp & Reference \\
		\hline
		RF++ & Stand-alone software (binary) & F & C, R & Yes (P) & \cite{karpievitch_introspective_2009} \\
    & (https://sourceforge.net/projects/rfpp) & & & & \\
		Historical RF & R package htree (CRAN) & F & C, R & Yes (P) & \cite{htreeR} \\
		MERT & R program (from Dr. Ahlem Hajjem) & T & R & No & \cite{hajjem_mixed_2011} \\
		MERF & R program (from Dr. Ahlem Hajjem) & F & R & No & \cite{hajjem_mixed-effects_2014} \\
		RE-EM & R package REEMtree (CRAN) & T & R & No & \cite{REEMtree, sela_re-em_2012} \\
		(S)MERT, (S)MERF & R package LongituRF (CRAN) & T, F & R & No & \cite{LongituRF, capitaine_2021} \\
		(S)REEMforest & & & & & \\
		GMERT & R code (supplement to original paper) & T & C & No & \cite{hajjem_generalized_2017}  \\
		BiMM forest & R code (supplement to original paper) & F & C & No & \cite{speiser_bimm_2019, speiser_bimm_2020} \\
		Multivariate RF & R package MultivariateRandomForest (CRAN) & F & R & No & \cite{MultivariateRandomForest, rahman_2017, segal_multivariate_2011} \\
		 & R package mvpart (CRAN) & T & R & No & \cite{death_multivariate_2002, mvpart} \\
		 & R package randomForestSRC (CRAN) & F & C, R & Yes & \cite{randomForestSRC, segal_multivariate_2011} \\
		\hline
		\multicolumn{6}{l}{VImp = Variable importance, T = tree, F = forest, R = regression, C = classification, P = permutation}
	\end{tabular}
}
\label{tab:implementation}
\end{table}

\section{Discussion}\label{sec:discussion}

In this paper, we review extensions of the CART-based random forest algorithm for the analysis of longitudinal data. Longitudinal data are common in areas such clinical trials and precision medicine. Using tree- or forest-based methods for analysis may help in patients stratification, disease progression prediction, and target biomarker identification for drug design. The repeated measurements of the same subject naturally induce clustering effect in longitudinal data, which negatively affects the variable selection performance and predictive accuracy of standard RF. To mitigate such effect, subject-level bootstrap re-sampling strategy can be considered. This approach works well for repeated measurements where observation ordering is not important. However, this approach does not take time effect into consideration, also there is no random effects included which limits its application for prediction. Historical RF is an approach which summarizes the observation histories and uses them along with concurrent observation to model the conditional mean of response variable. Extensions based on (generalized) linear mixed effects model, from both frequentist and Bayesian perspectives, provide another solutions to longitudinal structure, where tree/forest models are used to model the fixed effects component in the mixed effects model. Finally, by adjusting the splitting criterion in the construction of decision tree of RF, multivariate longitudinal response variables can also be handled.

In this review, we limit our focus on CART-based RF methods, there are certainly other approaches to be considered. Examples include GUIDE \cite{loh_regression_2002} and conditional inference \cite{hothorn_unbiased_2006} approaches. Some extensions reviewed here such as MERT and MERF can easily switch to their approaches for tree/forest construction, while others may require different extensions for longitudinal data. See \cite{loh2014fifty} for a review on different tree construction approaches and related discussions on extensions for longitudinal data analysis.

As we pointed out in Section \ref{sec:intro}, missing values represent a major challenge. Segal (1992) \cite{segal_tree-structured_1992} considered the surrogate splitting variable approach, which is one of the standard solutions for missing values in the literature of tree and forest methods. However, different missing mechanisms may require different approaches to handle missing values. In general, this is still an important research area in the context of developing and applying statistical methods and machine learning approaches in general.

Variable importance measure is a unique feature that RF can offer to support variable selection. Some reviewed extensions consider permutation-based variable importance which is easy to implement but computationally expensive. Other approaches such as the variable-delete approach may also considered, but correlation between predictor variables may negatively affect its performance. How to measure variable importance in a tailored fashion for longitudinal data warrents further study, because this could be beneficial in both understanding disease progression and searching for target biomarkers for drug design.

Another direction for future methodology development is on the effective handling of high dimensional longitudinal data. Except for the BiMM and REEMforest methods, the other reviewed methods have so far not been evaluated on high dimensional data sets. For instance, for the extensions from GLMM with non-quantitative outcomes, there is a need for an initial GLM fit which would be very difficult, if not impossible, with the high dimensional data sets. Omics data are usually high dimensional and may change over time. Having RF extensions being able to handle such data would provide fruitful insights on their effects on complex diseases.

Lastly, to the best of our knowlege, systematic benchmark studies to compare these aforementioned RF extensions have not been published so far. For instance, from the prediction perspective, how would the subject-level bootstrapping methods and historical RF compare with extensions with mixed effects model? Such benchmark studies would be informative and important for practitioners analyzing real data sets. This would be another important study to conduct.

\section*{Funding}

This work was supported by the German Federal Ministry of Education and Research (BMBF) funded e:Med Programme on systems medicine [grant 01ZX1510 (ComorbSysMed) to SSzy].

\bibliography{references}
\bibliographystyle{plain}

\end{document}